\documentclass[runningheads]{eccv_template/llncs}
\usepackage{graphicx}

\usepackage{tikz}
\usepackage{comment}
\usepackage{amsmath,amssymb} 
\usepackage{color}

\usepackage[accsupp]{axessibility}  

\usepackage{siunitx}
\usetikzlibrary{calc}
\usetikzlibrary{positioning}
\usetikzlibrary{spy}
\usetikzlibrary{shapes.geometric}
\usetikzlibrary{arrows.meta}
\usepackage{todonotes}
\usepackage{color}
\usepackage{colortbl}
\usepackage{tabularx}
\usepackage{graphicx} 
\usepackage[skip=0.333\baselineskip]{subcaption} 
\usepackage{float}
\usepackage{wrapfig}
\usepackage[hyphens,spaces,obeyspaces]{url}
\usepackage[colorlinks,allcolors=blue]{hyperref}
\usepackage{breakcites}
\usepackage{mathtools}
\DeclarePairedDelimiterX{\infdivx}[2]{(}{)}{%
 #1\;\delimsize\|\;#2%
}

\newcommand{\norm}[1]{\left\lVert #1 \right\rVert}

\begin{document}



\pagestyle{headings}
\mainmatter
\def\ECCVSubNumber{6112}  

\title{Neural Strands: Learning Hair Geometry \\and Appearance from Multi-View Images} 

\titlerunning{Neural Strands}
%



\author{Radu Alexandru Rosu\inst{1}\thanks{Work done during an internship at Reality Labs Research, Pittsburgh, PA, USA.}\index{Rosu,Radu Alexandru} \and
Shunsuke Saito\inst{3} \and
Ziyan Wang\inst{2,3} \and \\
Chenglei Wu\inst{3} \and 
Sven Behnke\inst{1} \and
Giljoo Nam\inst{3}
}

\authorrunning{Radu Alexandru Rosu et al.}
%
\institute{
University of Bonn, Germany
\and
Carnegie Mellon University
\and
Reality Labs Research
\\
\medskip
{\href[pdfnewwindow=true]{https://radualexandru.github.io/neural_strands/}{\nolinkurl{radualexandru.github.io/neural_strands}}}
}










\newcommand{\GN}[1]{{\color{red}{\bf GN: #1}}}
\newcommand{\ALEX}[1]{{\color{orange}{\bf ALEX: #1}}}
\newcommand{\SH}[1]{{\color{purple}{\bf SS: #1}}}
\newcommand{\ZW}[1]{{\color{red}{\bf ZW: #1}}}
\newcommand{\CW}[1]{{\color{cyan}{\bf CW: #1}}}

\newcommand{\NEW}[1]{{\color{black}{#1}}} 

\newcommand{\NEWNEW}[1]{#1} 

\newcommand{\NEWW}[1]{#1} 

\newcommand{\IGNORE}[1]{}
\newcommand{\etal}{~\textit{et al.}~}
\newcommand{\mparagraph}[1]{\noindent\textbf{#1.}}

\definecolor{ph-purple}{RGB}{129, 39, 232}
\definecolor{ph-blue}{RGB}{5, 131, 227}
\definecolor{ph-gray}{rgb}{0.5, 0.5, 0.5}
\definecolor{ph-orange}{RGB}{227, 127, 5}
\definecolor{ph-green}{RGB}{0, 135, 124}
\definecolor{ph-yellow}{RGB}{235, 201, 52}
\definecolor{ph-light-green}{RGB}{181, 209, 21}
\definecolor{ph-red}{RGB}{250, 101, 60}
\colorlet{ph-orange-light}{ph-orange!70}
\colorlet{ph-blue-light}{ph-blue!70}
\colorlet{ph-purple-light}{ph-purple!70}
\colorlet{ph-green-light}{ph-green!70}
\definecolor{ph-light-gray}{rgb}{0.75, 0.75, 0.75}

\newcommand{\reffig}[1]{Fig.~\ref{#1}}
\newcommand{\reftab}[1]{Tab.~\ref{#1}}
\newcommand{\refsec}[1]{Sec.~\ref{#1}}

\maketitle

\begin{abstract}
    We present Neural Strands, a novel learning framework for modeling accurate hair geometry and appearance from multi-view image inputs. The learned hair model can be rendered in real-time from any viewpoint with high-fidelity view-dependent effects. Our model achieves intuitive shape and style control unlike volumetric counterparts. To enable these properties, we propose a novel hair representation based on a neural scalp texture that encodes the geometry and appearance of individual strands at each texel location. Furthermore, we introduce a novel neural rendering framework based on rasterization of the learned hair strands. Our neural rendering is strand-accurate and anti-aliased, making the rendering view-consistent and photorealistic. Combining appearance with a multi-view geometric prior, we enable, for the first time, the joint learning of appearance and explicit hair geometry from a multi-view setup. We demonstrate the efficacy of our approach in terms of fidelity and efficiency for various hairstyles.

\end{abstract}

\section{Introduction}

Photorealistic rendering of digital humans plays an important role in many AR/VR applications such as virtual telepresence. 
In recent years, data-driven approaches have shown compelling results on geometry and appearance modeling of digital humans, especially for face~\cite{steve_meshvae,li2017flame,tewari2017mofa,tran2018nonlinear3dmm} and body~\cite{saito2019pifu,bagautdinov2021driving,xiang2020monoclothcap}.
Hair, on the other hand, still remains a challenge due to \NEW{the sheer number of thin hair strands}, their complex geometric structures, and non-trivial light transport effects such as subsurface scattering and specular reflections at microscale. 

\input{imgs_tex/teaser2.tex}

To enable strand-accurate hair reconstruction, a recent work~\cite{nam2019lmvs} leveraged explicit line assumption in multi-view stereo reconstruction. However, the reconstruction does not provide complete hair strands from the root on the scalp due to heavy self-occlusions. To date, connecting line segments from the scalp to the tip of hair for a variety of hairstyles remains difficult without strong data prior.

Appearance modeling of hair is also an active research field~\cite{khungurn2017azimuthal,benamira2021combined}. Physics-based rendering approaches typically require extensive light-transport computation to represent complex appearance of 3D hair strands, hence are too slow for real-time applications. Recently, data-driven approaches~\cite{chai2020neural,wei2018real} enable photorealistic rendering from geometric proxies such as orientation fields using neural rendering techniques. However, due to sub-optimal geometric quality and feature representations, these image-space neural rendering methods typically suffer from view-inconsistency and lack of fidelity. Volumetric rendering techniques~\cite{mildenhall2020nerf,steve_mvp}, on the other hand, achieve view-consistent novel-view rendering, but geometry-driven manipulation is not possible.


In this work, we present Neural Strands, a novel learning framework for jointly modeling hair geometry and appearance, which can be readily used for real-time rendering of photorealistic hair from an arbitrary viewpoint. Our idea is to build a strong data prior using a strand-based generative model learned from synthetic data. This allows us to register complete hair strands from the partial hair reconstruction obtained by~\cite{nam2019lmvs}. To parameterize the appearance and geometry of complete hairstyles from registration, we further present a novel hair representation called neural scalp textures, where each texel on a UV texture stores a feature vector describing both the shape and appearance of a single strand at a corresponding scalp position. With the aforementioned strand generator, the neural scalp texture is decoded into dense 3D strands, which are then rendered into RGB images by a neural renderer. 

\begin{figure*}
\centering
\begin{tikzpicture}

\newcommand\Shift{2.0}
\newcommand\TextY{-2.0}

\node[inner sep=0pt] at (\Shift*0,0)
    { \includegraphics[width=1.0\textwidth]{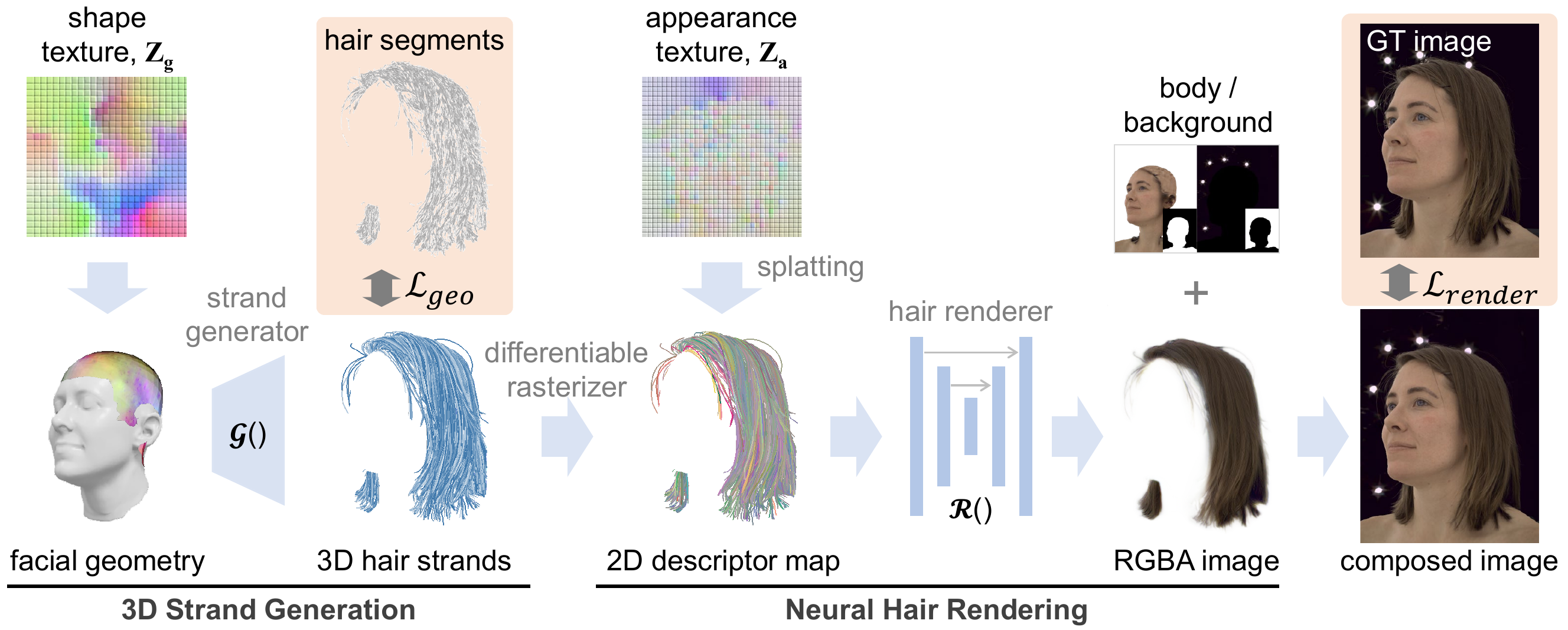} };

\end{tikzpicture}
\caption{
System overview. A neural scalp texture describing the strand shapes is embedded onto the scalp. A strand generator decodes the shape descriptors into explicit strands. The 3D strands are then rasterized to the screen space. Finally, a hair renderer decodes the 2D descriptors into an RGBA image of the hair which is composited with the body and background. We train our system end-to-end with both a geometrical loss towards sparse line segments and a rendering loss towards the ground-truth images.
} \label{fig:overview}
\end{figure*}

Our strand generator is a multilayer perceptron network that takes as input a strand shape feature vector and outputs the 3D shape of the strand. 
Inspired by neural ordinary differential equations~\cite{ode}, we design our strand generator to yield the first-order derivatives of a strand geometry, i.e., directional vectors with magnitudes. 
A complete 3D strand shape can be obtained by integrating the derivatives.
This formulation allows for generation of smooth strands and also enables a trade-off between compute and accuracy by changing the integration step size.
To learn a generic strand generator, we pre-train the network on a wide variety of strand data, resulting in a generic strand prior for robust strand fitting to the noisy real-world scan data. 
%

In order to render photo-realistic hair images using the generated strands, we propose a neural hair renderer.
The renderer comprises two parts: a differentiable rasterizer and an image synthesizer.
The rasterizer splats the appearance feature vectors onto 2D images while being differentiable w.r.t. the 3D positions of strands and the splatted features. 
The image synthesizer is a UNet architecture that takes as input the rasterized feature map and yields the final rendered images.
We train all the components in an end-to-end manner with direct image supervision. With an explicit strand representation, our method can directly edit or control the hair for rendering, e.g. for virtual haircut or hair blowing, which differentiates our approach from the recent work on free-viewpoint volumetric rendering~\cite{steve_nvs,steve_mvp,mildenhall2020nerf,park2021nerfies}.

%
In summary, our main contributions are:
\begin{itemize}
    \setlength\itemsep{0em}
    \item the first \IGNORE{end-to-end}\NEW{joint} learning framework for high-quality strand geometry and appearance from multi-view images, which can be readily used for photo-realistic real-time rendering of human hair,
    \item a highly expressive strand generative model, which enables strand-level hair registration from partial scan data,
    \item a novel neural scalp texture representation that compactly models explicit hair strand geometry and view-dependent appearance for real-time rendering
    \item an experiment showing that our approach enables intuitive manipulation of 3D hair and its rendering, which remains challenging for volumetric approaches.
\end{itemize}

\IGNORE{
Photorealistic rendering of digital humans plays a key role in many AR/VR applications such as virtual telepresence. In recent years, data-driven approaches have shown compelling results on geometry and appearance modeling of digital humans, especially for face~\cite{steve_meshvae,li2017flame,tewari2017mofa,tran2018nonlinear3dmm} and body~\cite{bagautdinov2021driving,xiang2020monoclothcap}. Hair, on the other hand, remains less explored due to its complex geometric structures and non-trivial light transport effects such as subsurface scattering and specular reflections at microscale. 

\input{imgs_tex/teaser.tex}
\SH{I would propose restructuring intro so that it's straightahead to the main point. 1st para: motivation+challenge in hair modeling, 2nd paragraph: intro to Neural Strands (stay high-level), 3rd: neural strands continue, geometric representation focused, 4th: rendering focused (contrast with volume), 5th: experiments focused, 6th: contribution}

To model 3D hair geometry, a strand-based representation has been widely used as it follows the geometric structure of actual hair strands. However, it is not trivial to apply traditional image-based modeling techniques, such as multi-view stereo (MVS), to get high quality 3D hair strands from images even from a dense set of multi-view images due to the aforementioned challenges. Nam \etal~\cite{nam2019lmvs} proposed a line-based multi-view stereo (LMVS) method to acquire 3D line segments of hair by employing the geometric structure of hair strands. The reconstructed hair strands are highly accurate, but they only reconstruct the outer surface of hair. Sun \etal~\cite{sun2021hairinverse} further improved LMVS by using multi-view photometric data and estimated hair reflectance properties for photo-realistic relighting. However, they rely on computationally heavy ray-tracing rendering and only estimate a single BSDF parameter for each subject. 

\begin{figure*}
\centering
\begin{tikzpicture}

\newcommand\Shift{2.0}
\newcommand\TextY{-2.0}

\node[inner sep=0pt] at (\Shift*0,0)
    { \includegraphics[width=1.0\textwidth]{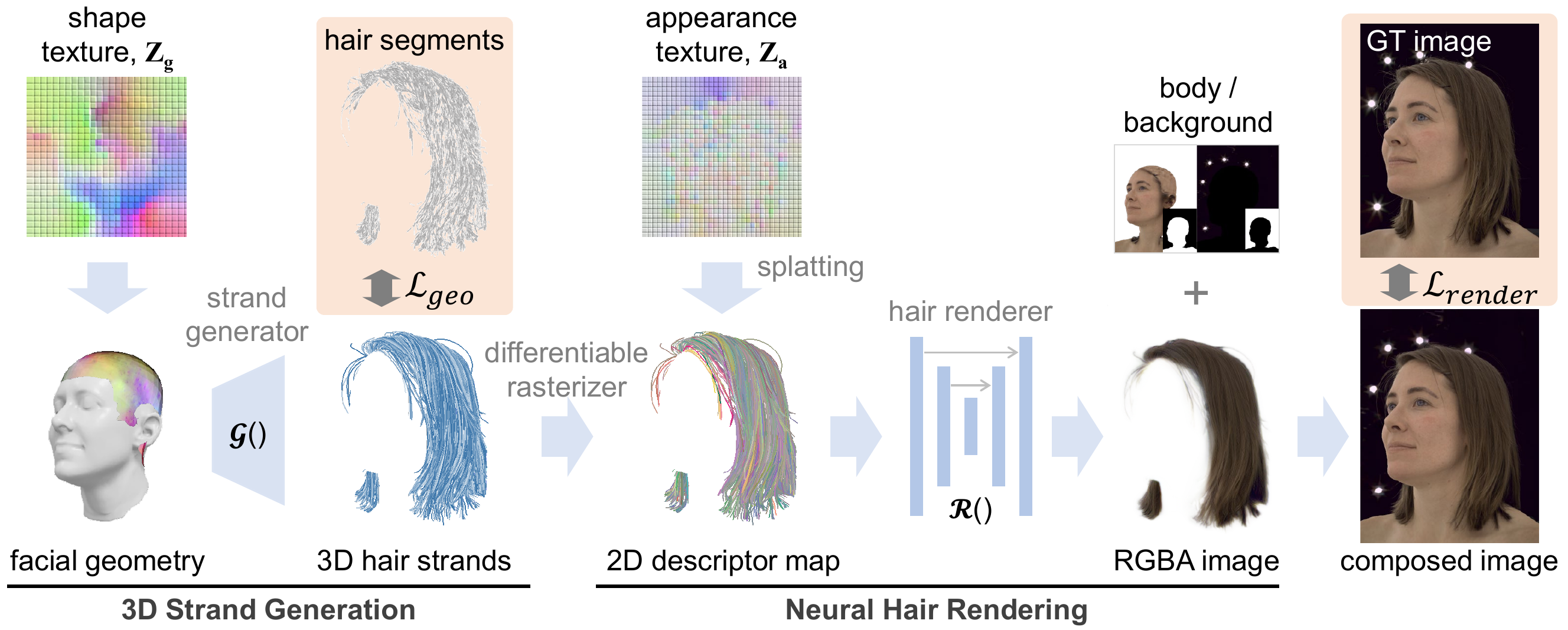} };

\end{tikzpicture}
\caption{
System overview. A neural scalp texture describing the strand shapes is embedded onto the scalp. A strand generator decodes the shape descriptors into explicit strands. The 3D strands are then rasterized to the screen space. Finally, a hair renderer decodes the 2D descriptors into an RGBA image of the hair which is composited with the body and background. We train our system end-to-end with both a geometrical loss towards sparse line segments and a rendering loss towards the ground-truth images.
} \label{fig:overview}
\end{figure*}

On the other hand, volumetric representations have been proposed as agnostic solutions for capturing hair geometry and appearance. Neural scene representations, such as Neural Volumes~\cite{steve_nvs} and NeRF~\cite{mildenhall2020nerf}, can capture hair geometry and appearance from multi-view images without hair specific priors. Recently, Lombardi \etal~\cite{steve_mvp} proposed the mixture of volumetric primitives (MVP) for efficient modeling and rendering of digital humans with realistic hair. While these works can represent high quality hair with details, they lack the ability to control or edit hair strands, which is often required in AR/VR applications (e.g., changing hairstyles). 

In this work, we present Neural Strands, a novel neural hair representation for photo-realistic modeling and rendering of human hair. Given multi-view images of a subject, Neural Strands recover high quality strand-based geometry together with an appearance model that generates realistic and multi-view-consistent renderings. Neural Strands are implemented with a neural scalp texture, a 2D scalp texture with feature vectors. The features stored on the scalp are used to decode both explicit strand geometry as a sequence of points and appearance. In order to fit explicit strands given multi-view data, we propose to use a strand generator that, given a latent representation, outputs the strand as a sequence of points. In order to constrain the problem, we pre-train the network on synthetic data and use sparse line-segments from multi-view stereo. Additionally, we train a neural renderer that, using strand features, recovers a full image of the hair. 

Neural Strands (the fitted geometry and appearance) can be used for animation, novel-view rendering and hair interpolation. We demonstrate our method with various real and synthetic multi-view captures and show that our method recovers both high quality strands and realistic appearance. Additionally, we manipulate the scalp texture to generate novel hair-styles not seen during training.

In summary, our main contributions are: 
\begin{itemize}
    \item A novel strand-based neural hair representation that captures explicit hair geometry and photo-realistic appearance.
    \item An end-to-end learning framework for stable and efficient training of the representation from multi-view images. 
    \item Demonstration of several applications such as novel hair-style generation and interpolation.
\end{itemize}


}

\section{Related Work}
\subsection{Image-based Hair Modeling}
\mparagraph{Strand-based Representation}
A common geometric representation for hair modeling is a set of 3D strands, where a strand is parameterized by a sequence of connected 3D points. 
To obtain 3D hair strands, typically multi-view hair capture methods first reconstruct a low-resolution base geometry using various image-based 3D modeling techniques, and run an additional strand-fitting process to obtain dense 3D hair strands that are connected to the scalp~\cite{beeler2012coupled,luo2012multi,luo2013structure,hu2014robust,hu2014capturing,hu2017simulation,herrera2012lighting}. 
However, these methods often lack fine details like flyaway hairs because of the low-resolution base geometry.
To overcome this limitation, Nam\etal\cite{nam2019lmvs} proposed a line-based multi-view stereo (LMVS) that is tailored to hair capture tasks and directly obtains 3D line segments from images. 
Sun\etal\cite{sun2021hairinverse} further improved LMVS and estimated the reflectance parameters of hair for photo-realistic rendering. 
However, the reconstructed strands from both methods~\cite{nam2019lmvs,sun2021hairinverse} only cover the outer surface of hair and they are not connected to the scalp. 
Another line of research focuses on single-view hair modeling~\cite{chai2012single,chai2013dynamic,chai2015high,chai2016autohair,zhang2017datadriven,zhou2018hairnet,zhang2019hair,yang2019dynamic}. 
While these works show promising results from less constrained capture setups, due to the ill-posed nature, these methods do not provide metrically accurate 3D geometry of hair strands. Moreover, they are not directly applicable to novel-view synthesis in a photorealistic manner. 
In contrast, our approach jointly learns metrically accurate geometry and appearance from multi-view images, enabling rendering from any viewpoint.  


\mparagraph{Volumetric Representation}
Volumetric representation is also used for hair modeling. 
Saito\etal\cite{saito20183d} propose to regress 3D hair from a single image using volumetric orientation fields as an intermediate representation, which can be easily handled by 3D convolutions.  
However, due to the low grid resolution of voxels, fine details such as flyaway hairs are not well represented. Combined with differentiable volumetric rendering, Neural Volumes~\cite{steve_nvs} and NeRF~\cite{mildenhall2020nerf} enable highly expressive geometry and appearance modeling of objects from multi-view images. 
Mixture of volumetric primitives (MVP) addresses the performance bottleneck of volumetric representations by representing scenes as a collection of small voxels~\cite{steve_mvp}. While these methods enable realistic rendering of 3D hairs, lack of geometric hair control hinders us from driving and intuitive manipulation of photorealistic hair models.


\subsection{Neural Hair Rendering}
Neural rendering~\cite{Tewari2020NeuralSTAR} has recently gained great attention for rendering photo-realistic images. Given 2D segmentation masks~\cite{tan2020michigan} and 2D orientation maps~\cite{olszewski2020intuitive,jo2019sc,qiu2019two,tan2020michigan}, generative adversarial networks (GANs) are trained to create RGB hair images that match the input data. By rendering these 2D features from 3D hair strands, these approaches allow us to photorealistically render hair images as well~\cite{wei2018real,chai2020neural}. However, we observe that rendering quality of these image-based approaches is highly dependent on the conditioned 2D features, and \IGNORE{the lack of fidelity in the underlying geometry}\NEW{often} leads to view-inconsistent results with limited fidelity. 
%
In this work, we show that highly accurate strands with a per-strand appearance code improve view consistency and fidelity of neural hair rendering.



\IGNORE{
\subsection{Hair representations}
Hair is challenging to reconstruct in 3D as it can have high-frequency detail, and self-occlusions which are difficult to model with traditional 3D meshes. Therefore, various representations have been developed in order to reconstruct hair, with different advantages and disadvantages.

\paragraph{Hair planes}
Hair planes represent the whole hair as a series of coarse polygons with a hair texture attached. Through the usage of a multitude of planes and alpha blending, good fidelity can be achieved. Hair planes are a common representation in video games, where latency is more important than fidelity. However, the usage of coarse polygonal shapes limits the representation to fairly simply hair-styles and high-frequency detail like stray hairs are usually ignored. 
This representation type is usually created by trained artists and not as output of an automated system.

\paragraph{Occupancy and flow}
A typical representation in deep learning is to disentangle the hair in and occupancy and flow field. The occupancy establishes the volume that the hair occupies and the flow describes how the hair direction inside the volume. Explicit strands are usually recovered in a post-processing step in which strands are grown according to the flow field. Since they are volumetric representations, they are processed with 3D network and therefore the fields are limited to relatively coarse resolutions.

\paragraph{Explicit strands}
Explicit hair strands as a series of 3D point can also be recovered like in the work of HairNet.

\paragraph{Implicit representation}
Recently, implicit representations have seen a surge in popularity with the work of NeRF. While not specifically designed for hair reconstruction, they can also represent this fine detail. The 3D structure is implicitly built inside the weights of the neural network. Rendering is done by querying the network densely inside the scanning volume and volumetric rendering is used to accumulate radiance and color. Further works like MVP, extend this idea in order to reduce the sampling space and increase the speed. 
Despite this, the representation is difficult to work with as it is not trivial to recover an explicit representation, animate or disentangle shape and color from implicit representations. Furthermore, the sub-millimeter detail of hair requires exceedingly small ray marching steps which slows down rendering.

\subsection{Hair rendering}
    GANs?
    Physically-based (the ones that fit a parametric BRDF or something like that) 
}

\section{Overview}
\label{sec:overview}
\reffig{fig:overview} shows the overview of our learning framework.
Neural Strands consists of three parts: neural scalp textures, strand generator, and neural hair renderer.
A neural scalp texture is a 2D UV-texture that stores either the shape (shape texture, $\mathbf{Z}_g$) or the appearance (appearance texture, $\mathbf{Z}_a$) of strands.
The strand generator, $\mathcal{G}()$, is a generative neural network that transforms a strand feature vector into a 3D strand geometry.
Finally, the hair renderer, $\mathcal{R}()$, is a UNet-architecture that renders hair images from rasterized appearance feature maps.

The design of Neural Strands is motivated by several hair-specific attributes.
First, in terms of geometry, each strand has its own attachment point in 3D scalp position.
By using an explicit scalp UV-map, we can easily control or edit the geometric features of strands based on the scalp location.
The shape texture $\mathbf{Z}_g$ exploits this property and enables several applications such as virtual haircut and hairstyle manipulation.
Second, while hair shows extremely complex appearance in 2D images, individual strands have a smooth color variation along their strand directions.
The complex hair appearance is determined by how the strands are shaped in 3D and how they are projected to each viewpoint. 
Therefore, we only store the low-frequency appearance information of each strand using our appearance texture $\mathbf{Z}_a$.
The high-frequency appearance in 2D images can be effectively represented by our rasterizer and hair renderer. 
This separation of per-strand appearance modeling and rendering is one of the keys that enable our high-fidelity results. 
In the following sections, we explain the details of each component and how they form the final hair images (Section~\ref{sec:neural_strands}) and then present the training procedure of our learning framework (Section~\ref{sec:train}). 

    
\section{Neural Strands}

\label{sec:neural_strands}



\subsection{Neural Scalp Textures}
Neural scalp textures are our base hair representation for explicit 3D geometry and appearance of strands.
Each texel in the shape texture $\mathbf{Z}_g$ stores a feature vector $\mathbf{z}_g^i \in \mathbb{R}^{D_g}$ that conveys the shape information of a single hair strand. Since the strand roots are created in 3D to ensure uniform coverage, 
the strands sample their corresponding $\mathbf{z}_g$ from the scalp texture using bilinear interpolation.
Similarly, the appearance texture is denoted as $\mathbf{Z}_a$ and $\mathbf{z}_a^i \in \mathbb{R}^{D_a}$ stores the appearance information of the strand. 
For the remaining of this paper, we assume that the scalp UV-mapping is known in advance.
We set the texture resolutions to $256^2$ for $\mathbf{Z}_g$, $512^2$ for $\mathbf{Z}_a$, and $D_g=64$, $D_a=16$. 

\begin{wrapfigure}{R}{0.5\textwidth}

\centering
\begin{tikzpicture} [scale=0.93,square/.style={regular polygon,regular polygon sides=4, minimum width = 20,
    minimum height = 20,
    inner sep = 0pt }]

\newcommand\Shift{1.2}
\newcommand\SpacingY{0.6}
\newcommand\TextY{-0.25}


\node[draw, square, fill=ph-red, rounded corners=1] at (\Shift*1, \SpacingY*2) {z};
\node[draw, square, fill=ph-red, rounded corners=1] at (\Shift*1, \SpacingY*1) {z};
\node[draw, square, fill=ph-red, rounded corners=1] at (\Shift*1, \SpacingY*0) {z};
\node[draw, square, fill=ph-red, rounded corners=1] at (\Shift*1, -\SpacingY*1) {z};
\node[draw, square, fill=ph-red, rounded corners=1] at (\Shift*1, -\SpacingY*2) {z};

\node[draw, square, fill=ph-blue!30, rounded corners=1] (t0) at (\Shift*1.43, \SpacingY*2) { \scriptsize	 0.0};
\node[draw, square, fill=ph-blue!50, rounded corners=1] (t1) at (\Shift*1.43, \SpacingY*1) { \scriptsize	 0.2};
\node[draw, square, fill=ph-blue!70, rounded corners=1] (t2) at (\Shift*1.43, \SpacingY*0) { \scriptsize 0.5};
\node[draw, square, fill=ph-blue!80, rounded corners=1] (t3) at (\Shift*1.43, -\SpacingY*1) {\scriptsize 0.7};
\node[draw, square, fill=ph-blue, rounded corners=1] (t4) at (\Shift*1.43, -\SpacingY*2) {\scriptsize 1.0};

\node[draw, fill=ph-light-green!30] (mlp0) at (\Shift*2.9, \SpacingY*2) {\scriptsize Strand Gen};
\node[draw, fill=ph-light-green!30] (mlp1) at (\Shift*2.9, \SpacingY*1) {\scriptsize Strand Gen};
\node[draw, fill=ph-light-green!30] (mlp2) at (\Shift*2.9, \SpacingY*0) {\scriptsize Strand Gen};
\node[draw, fill=ph-light-green!30] (mlp3) at (\Shift*2.9, -\SpacingY*1) {\scriptsize Strand Gen};
\node[draw, fill=ph-light-green!30] (mlp4) at (\Shift*2.9, -\SpacingY*2) {\scriptsize Strand Gen};

\draw[shorten >=1pt,shorten <=1pt,dashed,->,gray] (t0) -- (mlp0);
\draw[shorten >=1pt,shorten <=1pt,dashed,->,gray] (t1) -- (mlp1);
\draw[shorten >=1pt,shorten <=1pt,dashed,->,gray] (t2) -- (mlp2);
\draw[shorten >=1pt,shorten <=1pt,dashed,->,gray] (t3) -- (mlp3);
\draw[shorten >=1pt,shorten <=1pt,dashed,->,gray] (t4) -- (mlp4);

\coordinate (shiftS0) at (0, 0);
\coordinate (shiftS1) at (0.15, 0);
\coordinate (shiftS2) at (-0.1, 0);
\coordinate (shiftS3) at (-0.2, 0);
\coordinate (shiftS4) at (0.05, 0);
\coordinate (shiftS5) at (0.2, 0);
\node[] (i10) at (\Shift*4.3, \SpacingY*2 ){};
\node[] (i11) at (\Shift*4.3, \SpacingY*1){};
\node[] (i12) at (\Shift*4.3, \SpacingY*0){};
\node[] (i13) at (\Shift*4.3, -\SpacingY*1){};
\node[] (i14) at (\Shift*4.3, -\SpacingY*2){};
\node[] (i15) at (\Shift*4.3, -\SpacingY*3){};
\fill[ph-blue!70] (i10) circle (1.5pt);
\fill[ph-blue!70] (i11) circle (1.5pt);
\fill[ph-blue!70] (i12) circle (1.5pt);
\fill[ph-blue!70] (i13) circle (1.5pt);
\fill[ph-blue!70] (i14) circle (1.5pt);

\draw[shorten >=3pt,shorten <=1pt,dashed,->,gray] (mlp0) -- (i10);
\draw[shorten >=3pt,shorten <=1pt,dashed,->,gray] (mlp1) -- (i11);
\draw[shorten >=3pt,shorten <=1pt,dashed,->,gray] (mlp2) -- (i12);
\draw[shorten >=3pt,shorten <=1pt,dashed,->,gray] (mlp3) -- (i13);
\draw[shorten >=3pt,shorten <=1pt,dashed,->,gray] (mlp4) -- (i14);

\draw [shorten >=8, ->, -{Stealth}]( $(i10)$ ) -- ( $(i11) + (shiftS1)$ );
\draw [shorten >=8, ->, -{Stealth} ]( $(i11)$ ) -- ( $(i12) -(shiftS1) +(shiftS2) $ );
\draw [shorten >=8, ->, -{Stealth}]( $(i12)$ ) -- ( $(i13) -(shiftS2)+(shiftS3)$ );
\draw [shorten >=8, ->, -{Stealth}]( $(i13)$ ) -- ( $(i14) -(shiftS3)+(shiftS4)$ );
\draw [shorten >=8, ->, -{Stealth}]( $(i13)$ ) -- ( $(i14) -(shiftS3)+(shiftS4)$ );
\draw [shorten >=8, ->, -{Stealth}]( $(i14)$ ) -- ( $(i15) -(shiftS4)+(shiftS5)$ );

\draw[dashed,->,gray] (5.6, 0) -- (6.5,0);

\node[] (s10) at (\Shift*5.8, \SpacingY*2 ){};
\node[] (s11) at (\Shift*5.8+0.15, \SpacingY*1){};
\node[] (s12) at (\Shift*5.8-0.1, \SpacingY*0){};
\node[] (s13) at (\Shift*5.8-0.2, -\SpacingY*1){};
\node[] (s14) at (\Shift*5.8+0.05, -\SpacingY*2){};
\fill[ph-blue!70] (s10) circle (1.5pt);
\fill[ph-blue!70] (s11) circle (1.5pt);
\fill[ph-blue!70] (s12) circle (1.5pt);
\fill[ph-blue!70] (s13) circle (1.5pt);
\fill[ph-blue!70] (s14) circle (1.5pt);
\draw [gray] plot [] coordinates { (s10) (s11) (s12) (s13) (s14) };

\node at (\Shift*5.1, \TextY) {\scriptsize Integrate};

\end{tikzpicture}
\caption{The strand generator takes as input the shape descriptor $z$ and a parameter $t \in [0,1]$ indicating the position between root and tip. It outputs a direction vector for each section of the strand. The directions are integrated from root to tip in order to recover the explicit 3D positions of the strand nodes. } \label{fig:strand_gen}
\end{wrapfigure}
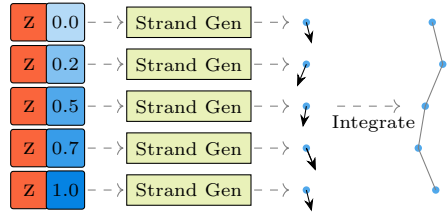

\subsection{Strand Generator}
A single hair strand $\mathbf{S}$ is a sequence of 3D points: $\mathbf{S}=\{ \mathbf{p}_k\}_{k=0}^L$, where $L$ is set to \num{100}. 
The full hair shape is a collection of strands $\{ \mathbf{S}^{j} \}$, where $j$ indexes over the strands.
Our strand generator $\mathcal{G}()$ is a neural network that transforms a shape feature vector $\mathbf{z}_g$ into a full 3D strand geometry $\mathbf{S}$, i.e., $\mathcal{G}():\mathbb{R}^{D_g} \rightarrow \mathbb{R}^{3 \times L}$.

\mparagraph{Network Architecture}
Inspired by NeuralODEs~\cite{ode}, we model the strand generation as an integration of hair gradients along each strand. 
We thus design the network to output finite differences along hair growing directions, i.e., $\mathbf{d}_k=\mathbf{p}_k - \mathbf{p}_{k+1}$, instead of 3D positions $\mathbf{p}_k$.

In order to model the high-frequency geometric details, we implement our strand generator using the modulated SIREN~\cite{modulatedsiren}.
Our strand generator has two multilayer perceptron (MLP) networks: a modulator and a synthesizer.
The modulator takes as input the strand shape feature $\mathbf{z}_g$ and outputs modulation vectors that modify the amplitude, frequency, and phase shift of the sinusoidal activation functions of the synthesizer network.
The SIREN\cite{siren}-based synthesizer takes as input a parameter $t \in [0,1]$ that indicates the relative position along the strand from root (0) to tip (1).
The output of the synthesizer is a 3D directional vector $\mathbf{d}_t$ with magnitude. 
The final positions of the strand nodes are obtained by the forward Euler method, i.e., $\mathbf{p}_k = \sum_{i=0}^{k} \mathbf{d}_i$.
See \reffig{fig:strand_gen} for illustration. 

\mparagraph{Discussion}
The benefit of working in the gradient domain is that each node of the strand has only local effect. 
When working directly with the positions, rotating the hair along the root node modifies the positions of all the subsequent nodes of the strand since they can all be considered as a kinematic chain. 
However, in the gradient domain, modifying the root node direction $\mathbf{d}_0$ does not necessarily modify the rest of the directions. 
This independence between nodes makes the learning easier, as the network does not need to learn the kinematic dependency between hair nodes. 
In addition, long straight hair can be easily represented by a network which outputs mostly constant directions. 

\subsection{Neural Hair Renderer}
Since we have an explicit hair geometry $\{ \mathbf{S}^j \}$ from our strand generator, we now render the hair appearance using a differentiable rasterization-based neural renderer. 
We take inspiration from the Neural Point-Based Graphics (NPBG)~\cite{npbg} and the Deferred Neural Rendering (DNR)~\cite{deferred} which use a point/mesh rasterizer.
Both NPBG and DNR use a geometry proxy (points or mesh) in order to carry a neural descriptor either with a texture map or at a per-point level. 
The neural descriptors are rasterized to a screen space and a UNet regresses the final color image. 

In our case, the geometry proxy is a 3D line segment, and the neural descriptors are given by the appearance texture $\mathbf{Z}_a$.
However, hair has various properties that need to be properly addressed.
First, hair is not opaque and can show complex scattering effects. 
Second, the thin geometry also tends to create aliasing artifacts when rendered. 
We effectively solve these issues by using alpha blending and let the UNet renderer directly output the alpha maps for natural strand color blending. 

\mparagraph{Neural Hair Descriptor}
The appearance texture $\mathbf{Z}_a$ stores the vectors $\mathbf{z}_a^i$ for each texel position $i$.
For each strand $\mathbf{S}^{j}$, we bi-linearly interpolate the neighbouring four texels at the root to obtain the corresponding $\mathbf{z}_a$ for the strand.
The 3D points $\{ \mathbf{p}_k\}_{k=0}^L$ belonging to the same strand $\mathbf{S}^{j}$ share the same per-strand feature vector $\mathbf{z}_a$. 
A per-point neural descriptor is then defined as a concatenated vector of the strand feature, the point direction $\mathbf{d}$, and the $t$ parameter:
\begin{equation}
    \mathbf{g}=\left[ \mathbf{z}_a, \mathbf{d}, t  \right], \quad \mathbf{g} \in \mathbb{R}^{D_a+3+1}.
\end{equation}
The per-point descriptors are rasterized to each viewpoint via a differentiable line rasterizer.

\mparagraph{Differentiable Rasterization} 
\NEW{We project the neural descriptors to screen-space by rasterizing the strand lines. However, as na\"ive rasterization methods are non-differentiable, we replace the hard rasterizer with soft-rasterization~\cite{adop,soft_rast,differentiable_splatting}.
We first hard-rasterize unique strand indices onto the screen using OpenGL which allow to recover for each pixel a 3D point that lies on a non-occluded strand line. The 3D point is associated with a descriptor $\mathbf{\tilde{g}}$ by linearly interpolating the descriptors from the two points that define the strand segment. At the end of this step, we obtain a point cloud of the visible points of the hair together with their neural descriptors. 
As a second step, we project the cloud to screen and the descriptors are bi-linearly splatted to the neighboring pixels. 
In the case where multiple 3D points contribute to the same pixel, the splatted descriptor at the pixel is defined as the weighted average of the contributing 3D points:
\begin{equation}
    \mathbf{h}_u = \frac{ \sum w \cdot \mathbf{\tilde{g}} }{ \sum w},
\end{equation}
where the subscript $u$ is the pixel index in the rasterized image space, $\mathbf{h}_u$ is the rasterized and averaged descriptor at the pixel, and $w$ is the contribution weight of each descriptor $\mathbf{\tilde{g}}$.
This soft-rasterization for hair is crucial to ensure the rendering loss to be back-propagated to the neural scalp textures $\mathbf{Z}_g$ and $\mathbf{Z}_a$ as in Fig.~\ref{fig:overview}.
In order to deal with possible holes in the hair, we also splat at multiple resolutions and concatenate each resolution with the corresponding layer in the UNet, similar to the previous works~\cite{npbg,adop}.
}

\mparagraph{Image Generation} 
The multi-resolution descriptor maps are concatenated with per-pixel viewing directions in order to model view-dependent effects and are given as input to the UNet which predicts an RGB and an opacity map for the hair.
Effectively, the input to the UNet is a descriptor map of $(D_a+3+1+3)$ channels \NEW{concatenated to each downsampling stage of the network}, and the output is a four-channel image of RGB and alpha. 

We find that the intermediate activation maps of the UNet were aliased by the down-sampling with strided convolutions. 
Therefore, we replace the down-sampling and up-sampling layers of the UNet with the anti-aliased versions used in~\cite{karras2021alias} and the activation function with their filtered leaky ReLU. 
We find this change effectively solves the aliasing issue and removes temporal flickering artifacts when rendering novel-view images.


\mparagraph{Image Composition}
In order to blend the hair with the background and body parts, we also learn a low-resolution texture for a body mesh and a background mesh represented as a sphere around the subject.
The background, body, and hair are alpha-blended together in order to recover the full image. 
The compositing can be viewed in~\reffig{fig:overview}. 

\section{Training}
\label{sec:train}
Training Neural Strands is twofold. 
First, we pre-train the strand generator $\mathcal{G}()$ using synthetic hair models and freeze the parameters. 
Then, for each subject, we optimize the feature vectors of neural scalp textures $\mathbf{Z}_g$ and $\mathbf{Z}_a$, as well as the parameters of the UNet renderer $\mathcal{R}()$ using the pre-trained generator $\mathcal{G}()$. 

\subsection{Strand Generator}
\noindent\textbf{VAE Training.} 
The role of our strand generator $\mathcal{G}()$ is to provide a strong prior of realistic hair strand shapes that can be readily used for our image-based hair modeling framework. 
\NEWW{To this end, we train it in an auto-encoder fashion with synthetic 3D curves.}
Concretely, we implement it as a variational autoencoder (VAE)~\cite{vae} in order to obtain a smooth embedding of $\mathbf{z}_g$.
The input and output of the VAE is a strand, i.e., $\{\mathbf{p}_k\}_{k=0}^L$ .

We design a simple encoder network with a 1D CNN.
Given the 3D points of a strand, the encoder outputs the parameters $\mathbf{s}_\mu$ and $\mathbf{s}_\sigma$ of the Gaussian distribution over the latent variables. 
During training, we sample $\mathbf{z}_g$ from this distribution using the reparameterization trick:
$\mathbf{z}_g= \mathbf{s}_\mu +\epsilon \cdot \mathbf{s}_\sigma, \epsilon \sim \mathcal{N}(0,1)$.
Given the strand embedding $\mathbf{z}_g$, we decode it back to the original points using the decoder $\mathcal{G}()$ which is implemented as a modulated SIREN.

We use the L2 loss between the predicted and ground-truth 3D points. Since this loss gives little regard to high-frequency detail like curls, we add a loss on the predicted directions. The data term is defined as
\begin{equation}
    \mathcal{L}_{data} = \sum_{i=0}^{L}  \norm{ \mathbf{p}_i- \tilde{ \mathbf{p} }_i }_{2}^{2} + \lambda_{d} \left(1- \mathbf{d}_i \cdot \tilde{ \mathbf{d} }_i \right),
\end{equation}
where $\mathbf{p}$ and $\mathbf{\tilde{p}}$ are the original and reconstructed points, and $\mathbf{d}$ and $\mathbf{\tilde{d}}$ are their directions.
$\lambda_{d}$ is set to \num{1e-3}.
We also add the Kullback–Leibler divergence term $\mathcal{L}_{KL}$~\cite{vae}.
We train the VAE with the total loss:
\begin{equation}
    \mathcal{L}_{VAE} = \mathcal{L}_{data} 
        + \lambda_{KL} \mathcal{L}_{KL} \left(  \mathcal{N}(\mathbf{s}_\mu, \mathbf{s}_\sigma) \mid\mid \mathcal{N}(\mathbf{0},\mathbf{I}) \right), 
\end{equation}
where $\lambda_{KL}$ is set to \num{1e-3}.
Once the VAE is trained, we discard the encoder network and only use the decoder as our pre-trained strand generator $\mathcal{G}()$.

\mparagraph{Training Data}
\NEWW{The dataset to train the strand generator is a set of synthetic 3D curves and each curve represents a hair strand as a sequence of \num{100} points.}
To remove the variance between the strands, we represent each one in a local coordinate system defined by the root position and the tangent-bitangent-normal (TBN) at the scalp. 
We also augment each strand by randomly stretching each dimension, mirroring along the tangent and bitangent vectors and rotating along the normal.

\subsection{End-to-End Optimization}
\noindent\textbf{Data Preparation.} Given multi-view images as input, we first perform multi-view stereo to obtain 3D geometry of the subject.
We then fit the reconstructed face geometry to the FLAME face template~\cite{li2017flame}. 
This fitting process gives us a known UV-mapping for the scalp region and also effectively removes the hair geometry in the reconstructed mesh.
We also perform the line-based multi-view stereo (LVMS)~\cite{nam2019lmvs} to get partial hair strand reconstruction. 
Note that the partial strands only include the strands in the outer surface of hair and are not connected to the scalp. We additionally perform a diffusion algorithm based on user strokes similar to~\cite{chai2013dynamic} in order to resolve the directional ambiguity of the line segments and obtain a consistent direction of growth. 

\mparagraph{End-to-end training} 
The input to our end-to-end optimization framework are 1) multi-view images, 2) the fitted facial geometry, and 3) the partial hair strands. 
Given the pre-trained strand generator $\mathcal{G}$ and input data, we jointly optimize for the neural scalp textures $\mathbf{Z}_g$ and $\mathbf{Z}_a$ as well as the parameters of the UNet renderer $\mathcal{R}()$ for each captured subject.

\mparagraph{Geometric Loss}
The geometric loss encourages our strand generator $\mathcal{G}()$ to output hair strands that align with the partial hair strands from the LMVS~\cite{nam2019lmvs}. 
The loss is defined as the bi-directional Chamfer of the distance and directions between the two point clouds:
\begin{equation}
\begin{aligned}
    \mathcal{L}_{geo} = 
        \sum_{\mathbf{x} \in X} \Big(   {\left \| \mathbf{x}-\mathbf{y_x} \right \|}_2  
         +  \left( 1 - \mathbf{d}_x \cdot \mathbf{d}_y  \right)  \Big)  + 
        \sum_{\mathbf{y} \in Y} \Big(  {\left \| \mathbf{x_y}-\mathbf{y} \right \|}_2 
         + \left( 1 - \mathbf{d}_x \cdot \mathbf{d}_y  \right)  \Big) ,
\end{aligned}
\end{equation}
where $X$ is a set of 3D points in the generated strands, $Y$ is a set of points in the LMVS-reconstructed hair segments, and $\mathbf{y_x}$ represents the point $\mathbf{y}$ which is the closest one in $Y$ to the point $\mathbf{x}$. 
Effectively, the Chamfer distance brings the closest points closer together and also aligns their directions.

While the geometric loss alone gives us plausible hair geometry, it can lead to missing or sub-optimal fitting results as the LMVS-reconstructed hair segments do not cover the entire region of the hair. 
Because the strands that are not reconstructed from the LMVS could still be visible from the images, there is an opportunity to supervise the strand generation with the rendering loss. 
We therefore also use the rendering loss to optimize for the shape texture $\mathbf{Z}_g$. 
This is done by back-propagating the rendering loss not only to the appearance texture $\mathbf{Z}_a$, but also to the shape texture $\mathbf{Z}_g$.

\mparagraph{Rendering Loss}
The neural renderer $\mathcal{R}()$ together with the appearance texture $\mathbf{Z}_a$ is trained using a combination of L2 and LPIPS~\cite{lpips} loss. 
The rendering loss $\mathcal{L}_{render}$ is thus defined as:
\begin{equation}
    \mathcal{L}_{render} = 
    \sum_{n=1}^{N}
    \left ( 
        \norm{I_{n} - \tilde{I}_{n}}^2_2  
        + \lambda_{L} \mathcal{L}_{LPIPS}(I_{n}, \tilde{I}_{n})
    \right ),
\end{equation}
where $I_{n}$ and $\tilde{I}_{n}$ are the rendered and captured images from $n$-th view, and $N$ is the number of multi-view images.
We set $\lambda_{L}=0.1$.

\begin{figure}[t]
\centering

\def\hairAlpha{./imgs/alpha_prediction2/pred_alpha.png} 
\newlength{\WAL}
\newlength{\HAL}
\settowidth{\WAL}{\includegraphics{\hairAlpha}}
\settoheight{\HAL}{\includegraphics{\hairAlpha}}

\captionsetup[subfigure]{justification=centering,font=small,labelfont=small}

  \subcaptionbox{LMVS mask\label{fig:a}}%
  {\includegraphics[ trim=.0\WAL{} .1\HAL{} .1\WAL{} .0\HAL{},clip,  width=0.22\linewidth]{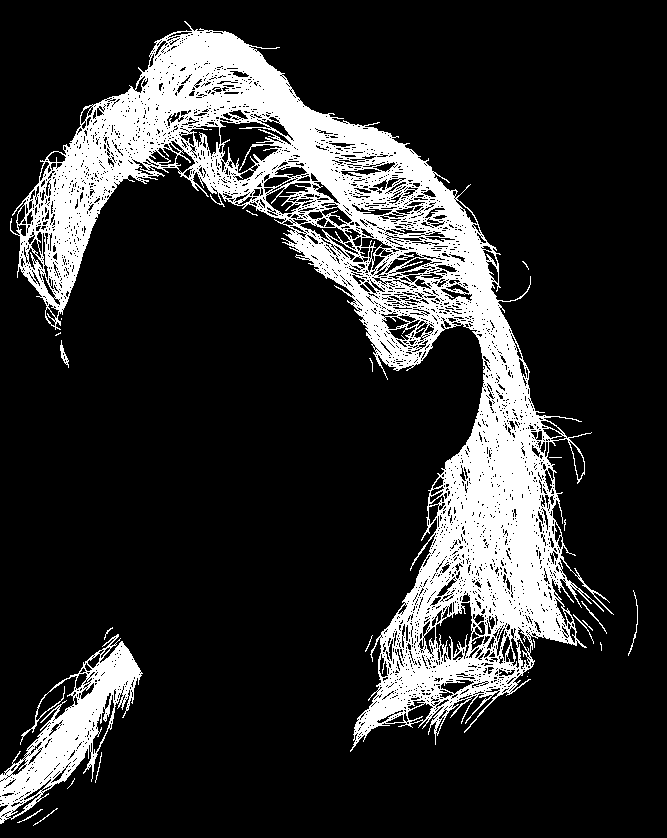}}
  \subcaptionbox{Mask trimap\label{fig:b}}%
  {\includegraphics[trim=.0\WAL{} .1\HAL{} .1\WAL{} .0\HAL{},clip, width=0.22\linewidth]{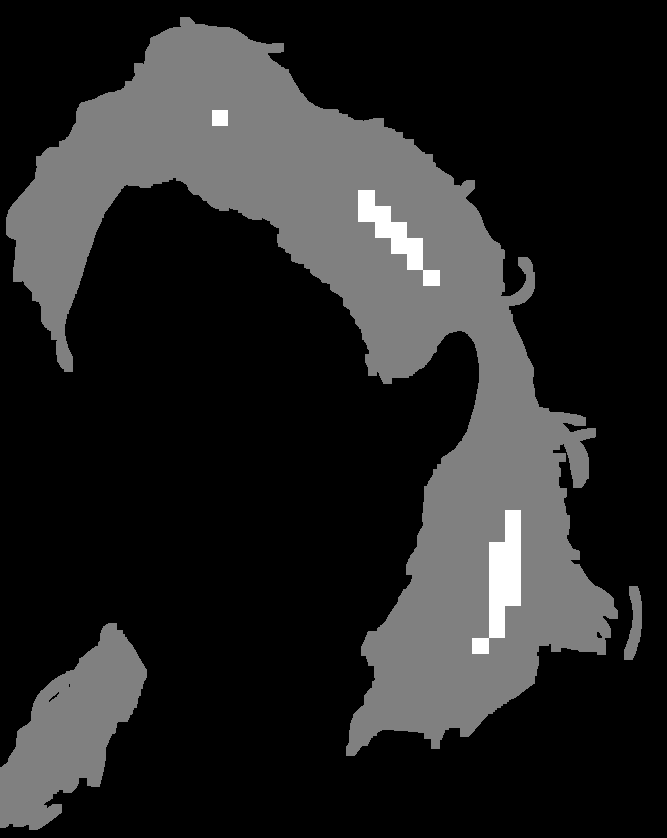}}
  \subcaptionbox{Predicted alpha\label{fig:c}}%
  {\includegraphics[trim=.0\WAL{} .1\HAL{} .1\WAL{} .0\HAL{},clip, width=0.22\linewidth]{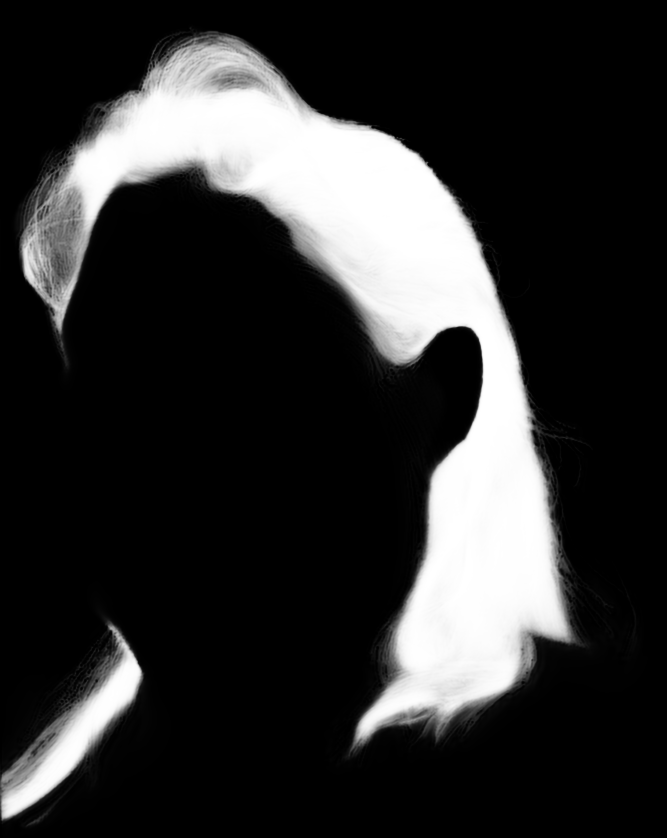}}
  \subcaptionbox{Novel image\label{fig:a}}%
  {\includegraphics[trim=.0\WAL{} .1\HAL{} .1\WAL{} .0\HAL{},clip, width=0.22\linewidth]{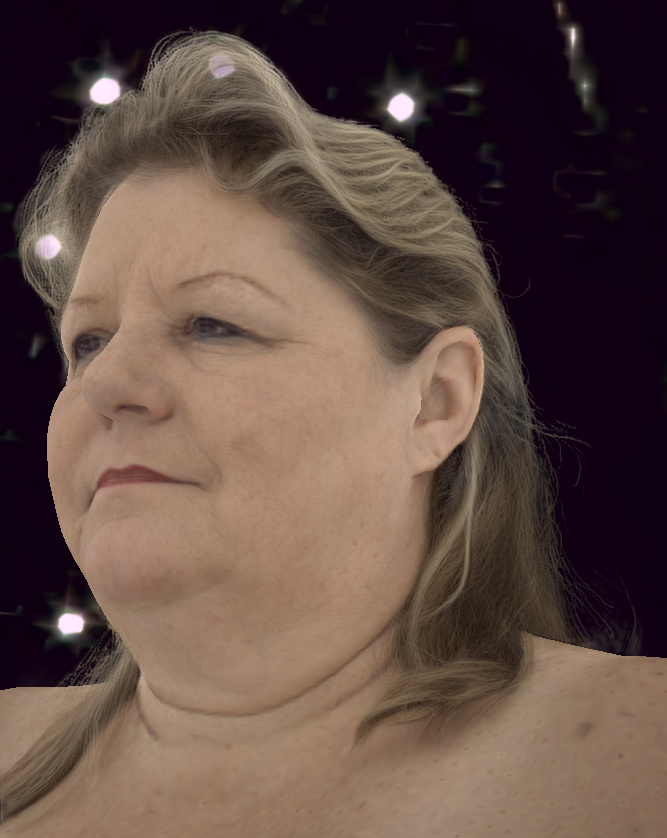}}

  \caption{ Alpha prediction and blending. (a) Mask from LMVS; (b) Trimap obtained from (a); (c) Predicted alpha map; (d) Composed image using (c).
  }\label{fig:alpha_pred}
\end{figure}

\mparagraph{Alpha Loss}
In order to also offer supervision to the predicted alpha, we rasterize the LMVS line-segments to the image. 
By itself, this hard LMVS mask would be inadequate to be used as alpha supervision since it enforces a strictly opaque hair. 
\NEW{To remedy this, we dilate the mask and define a region without hair that we can claim should be empty and therefore should have an alpha of 0.
After the dilation, we erode to define an interior region that we can be certain that it should be opaque. }
These two regions are used for supervision, while the border regions containing stray hairs are left unsupervised as we cannot reliably supervise their soft opacity. 
The alpha map loss $\mathcal{L}_{alpha}$ is defined as:
\begin{equation}
    \mathcal{L}_{alpha} = 
        \sum_{n=1}^{N}
            \left ( 
                \norm{A_{n} - \tilde{A}_{n}  }^2_2 \cdot M_{n}
            \right ),
\end{equation}
where $A_n$ and $\tilde{A}_n$ are the reference alpha map from LMVS and the generated alpha map from $n$-th view, and $M_n$ is their mask of union of the interior and exterior regions. 
See \reffig{fig:alpha_pred} for illustration.

\mparagraph{Total Loss}
The total loss for our end-to-end optimization is:
\begin{equation}
    \mathcal{L}_{total} = 
    \lambda_1 \mathcal{L}_{geo} 
    + \lambda_2 \mathcal{L}_{render} 
    + \lambda_3 \mathcal{L}_{alpha},
\end{equation}
where $\lambda_1$,$\lambda_2$, and $\lambda_3$ are set to \num{1}, \num{1e-3}, and \num{1e-3}, respectively.

\mparagraph{Optimization Details}
Fitting the geometry of the strands based on the Chamfer loss to the line-segments is a highly ill-posed problem.
To solve this, we propose two solutions: a coarse-to-fine and a root-to-tip optimization of the shape texture $\mathbf{Z}_g$. 
%
Coarse-to-fine optimization of the scalp imposes a smoothness prior on the features.
We implement this by sampling from the texture map in the forward pass and blurring the gradient of the loss w.r.t to the texture $ \frac{\nabla \mathcal{L}}{ \nabla \mathbf{Z}_{g}}$ in the backward pass, so that neighboring pixels receive similar gradients. 
We start by blurring with a large kernel and gradually decrease it until the gradient is propagated only towards the pixels that correspond to the strand root. 
This optimization scheme is similar to the Laplacian Pyramid from \cite{deferred}. 
However, instead of optimizing various textures at multiple resolutions, we optimize only one, which makes it faster for training and inference.

Root-to-tip optimization is performed by starting the training with only the roots of the strands and masking out the gradient for the rest of the strand vertices. 
We linearly anneal the rest of the nodes gradually during optimization. 
%
In addition, for stable training, we set $\lambda_2$ and $\lambda_3$ to \num{0} for the first 1,000 iterations since at the beginning, the strands are far away from the correct hair region in image-space.
In~\reffig{fig:coarse2fine} we show the impact of our coarse-to-fine and root-to-tip optimization scheme.

\section{Results}
    \begin{table}[t]
\parbox{.45\linewidth}{
        \begin{center}
    		\includegraphics[width=0.49\textwidth]{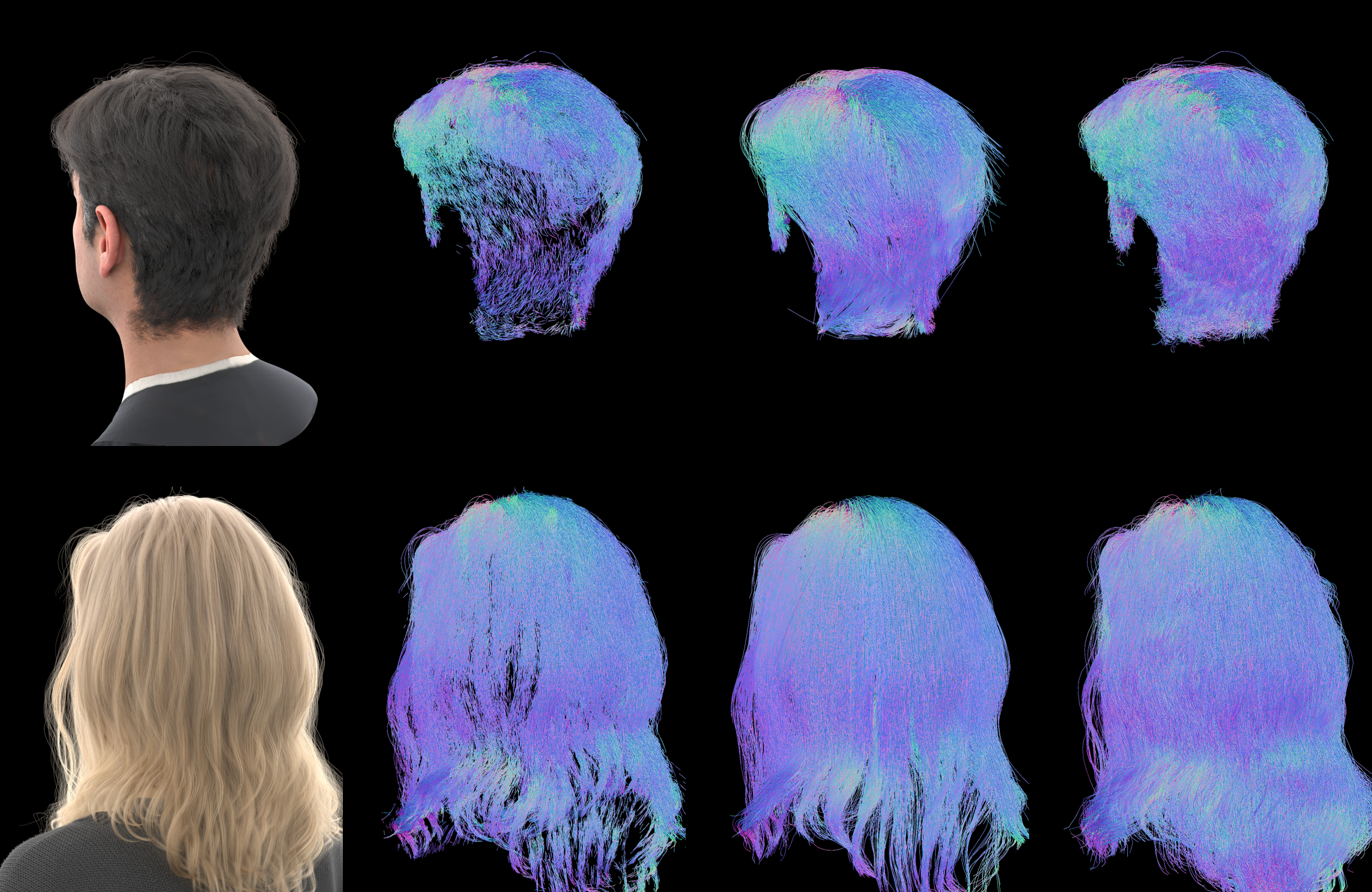}
    	\end{center}
    	\captionof{figure}{Synthetic data and geometry reconstructions. From left: GT image, LMVS geometry, our geometry, GT geometry. Note that the strands from LMVS are segmented and not connected to the scalp.}
    	\label{fig:synthetic}
}
\hfill
\parbox{.45\linewidth}{
        \begin{center}

    		\def\haircf{./imgs/coarse2fine/no_coarse2fine.png} 
            \newlength{\WCF}
            \newlength{\HCF}
            \settowidth{\WCF}{\includegraphics{\haircf}}
            \settoheight{\HCF}{\includegraphics{\haircf}}
            
              
              \begin{tikzpicture}  
                \node[inner sep=0pt] (noc2f) at (0.0, 0.0)
                {\includegraphics[  trim=.0\WCF{} .15\HCF{} .0\WCF{} .11\HCF{},clip,   width=0.47\linewidth]{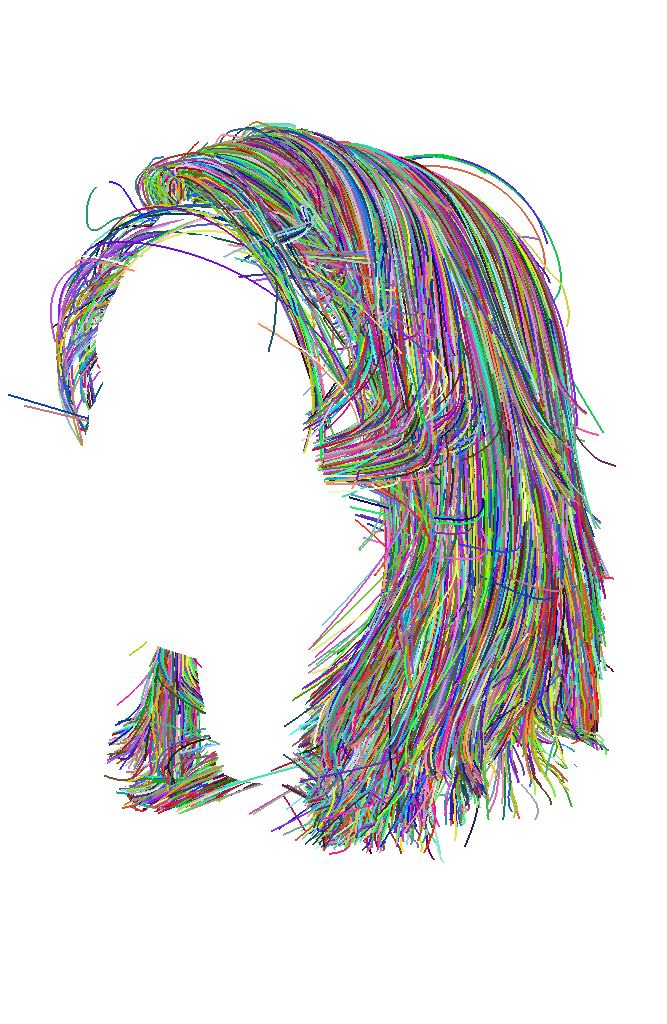}};
                \node[inner sep=0pt] (noc2f) at (2.7, 0.0)
                {\includegraphics[  trim=.0\WCF{} .15\HCF{} .0\WCF{} .11\HCF{},clip,   width=0.47\linewidth]{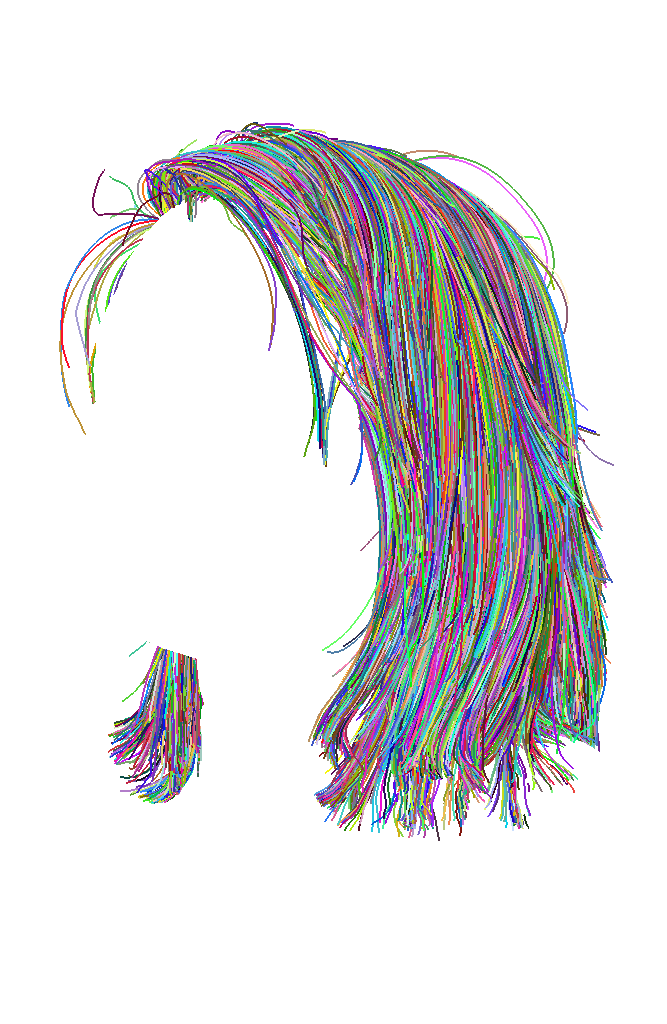}};
                \node at (0.0, -2.0) [text width=2.5cm] {\scriptsize a) No coarse-to-fine nor root-to-tip};
                \node at (2.7, -2.0) [text width=2.5cm]  {\scriptsize b) With coarse-to-fine and root-to-tip};
              \end{tikzpicture}

    	\end{center}
    	\captionof{figure}{Our coarse-to-fine and root-to-tip optimization of the shape texture helps the network converge to the correct shape.}
    	\label{fig:coarse2fine}
}
\end{table}%

We evaluate our method using both real and synthetic data.
For real images, we use a multi-view camera dome with $\sim$140 cameras uniformly distributed on a sphere of two meter diameter. 
For synthetic images, we use artist-created 3D hair models.
Virtual cameras are placed to mimic the real capture setup.
\reffig{fig:synthetic} shows the synthetic renderings of two 3D models with short and long hairstyles.
\NEW{We train our model for \SI{48}{\hour} on a single NVIDIA V100 GPU.}


\begin{table*}[!t]
    \scriptsize
    \centering
    \caption{Comparison between the previous work LMVS~\cite{nam2019lmvs} and our method using synthetic dataset.} 
    \begin{tabular}{|c|cccccc|cccccc|}
    \hline
     & \multicolumn{6}{c|}{Short hair}
     & \multicolumn{6}{c|}{Long hair}\\\hline
     $\tau_p/\tau_d$
     & \multicolumn{2}{c}{1mm / $10^\circ$}
     & \multicolumn{2}{c}{2mm / $20^\circ$}
     & \multicolumn{2}{c|}{3mm / $30^\circ$}
     & \multicolumn{2}{c}{1mm / $10^\circ$}
     & \multicolumn{2}{c}{2mm / $20^\circ$}
     & \multicolumn{2}{c|}{3mm / $30^\circ$}\\\hline
     Method
     & {\scriptsize LMVS} & {\scriptsize Ours }
     & {\scriptsize LMVS} & {\scriptsize Ours }
     & {\scriptsize LMVS} & {\scriptsize Ours }
     & {\scriptsize LMVS} & {\scriptsize Ours }
     & {\scriptsize LMVS} & {\scriptsize Ours }
     & {\scriptsize LMVS} & {\scriptsize Ours }\\\hline
     Precision
     & \textbf{56.91} & 52.79 & \textbf{93.42} & 92.94 & \textbf{98.85} & 98.18
     & 26.25 & \textbf{32.59} & \textbf{75.13} & 71.40 & \textbf{93.51} & 71.40\\
     Recall
     & 12.11 & \textbf{13.78} & 30.29 & \textbf{48.38} & 46.62 & \textbf{71.51}
     & \textbf{16.54} & 14.62 & 39.12 & \textbf{42.06} & 54.01 & \textbf{62.91}\\
     F-score
     & 19.98 & \textbf{21.85} & 45.75 & \textbf{63.64} & 63.36 & \textbf{82.75}
     & \textbf{20.30} & 20.19 & 51.45 & \textbf{52.94} & \textbf{68.47} & 66.89\\\hline
    
    \end{tabular}
    
    \label{tab:geo}
\end{table*}

\subsection{Evaluation with Synthetic Data}


Since it is impossible to obtain the ground truth geometry of hair strands from real captured images, we use synthetic data for the evaluation. 
In \reftab{tab:geo}, we show quantitative comparison on recovered strand geometries over the state-of-the-art hair geometry reconstruction method~\cite{nam2019lmvs}.
We follow the error metric from~\cite{nam2019lmvs,sun2021hairinverse} and show the precision, recall, and F-scores of the reconstructed 3D point clouds over their ground truth with various threshold values. 
It is shown that the previous work~\cite{nam2019lmvs} tends to have better precision (accuracy), whereas our method has better recall (completeness) and F-score values in general.
This shows the effectiveness of our method to reconstruct complete hair strands that are connected to the scalp.
\NEW{
We further emphasize that LMVS only recovers disjoint strand segments while our method recovers full strands of hair which enables further applications such as animations.
\reffig{fig:synthetic} illustrates the limitation of \cite{nam2019lmvs} and how Neural Strands overcomes it.
}

\subsection{Evaluation with Real Data}
We compare our method with two view-synthesis methods, NeRF~\cite{mildenhall2020nerf} and MVP~\cite{steve_mvp}, that can model and render hair appearance from captured multi-view images.
As shown in \reffig{fig:qualitative_compare}, our method is capable of rendering highly detailed hair textures, which is difficult to achieve with the other methods.

\reftab{tab:quantitative_compare_psnr} shows quantitative comparisons.
We compute PSNR, SSIM~\cite{ssim}, and LPIPS~\cite{lpips} for the hair region of nine novel-view images that are not used in training. 
The numbers show the averaged values of six subjects for each method.
While our method shows the best LPIPS loss with better visual quality, PSNR and SSIM values are slightly lower than the other methods.
This is also reported in previous works~\cite{lpips,nilsson2020,patel2019} on image quality metric; PSNR and SSIM do not properly reflect the perceptual quality of reconstructed images.
We also compare the rendering time of each method in \reftab{tab:quantitative_compare_time}. 
Note that we only need to run the decode step (strand generation) once, and the generated strands can be rendered to novel viewpoints in less than 40 ms, thus achieving real-time rendering of $>25$ frames per second. \NEW{All experiments were conducted on a NVIDIA V100 GPU}.

\input{imgs_tex/qualitative_compare.tex}
\begin{table}[!t]
\parbox{.45\linewidth}{
\centering
    \captionsetup[subfigure]{font=small,labelfont=small}
    \caption{
    We perform better under the perceptual metric (LPIPS), indicating that ours have more realistic looking hairs.}
    \label{tab:quantitative_compare_psnr}
    \begin{tabular}{r   | c c c }
     & \shortstack{NeRF} & \shortstack{MVP} & \shortstack{Ours} \\
     \hline
     PSNR ($\uparrow$)   &  31.71  &  \textbf{32.82} &  31.30\\
     SSIM ($\uparrow$)   &  0.9383 &  \textbf{0.9599} &  0.9452\\
     LPIPS ($\downarrow$)   &  0.1598 &  0.1226 &  \textbf{0.0811}\\
     \end{tabular}
}
\hfill
\parbox{.45\linewidth}{
\centering
    \captionsetup[subfigure]{font=small,labelfont=small}
    \caption{
    Our method can render a static hair in real-time ($>$25 fps) and dynamic strands at interactive rates ($>$13 fps).}
    \label{tab:quantitative_compare_time}
    \begin{tabular}{r   | c c c }
     & \shortstack{NeRF} & \shortstack{MVP} & \shortstack{Ours} \\
     \hline
     decode ($\downarrow$)   & -  & \textbf{20.40} & 34.54\\
     render ($\downarrow$)   & -  & 81.60 & \textbf{38.82} \\
     total (ms) ($\downarrow$)   & 27,910 &  102.00 &  \textbf{73.36}\\
     \end{tabular}
}
\end{table}
\input{imgs_tex/hair_manipulation.tex}

\subsection{Applications}
\NEW{In this section, we describe two demos that show the ability of post-capture manipulation of hair strands, which differentiates our method from other view-synthesis methods. Please see the supplemental video for better visualization.}

\mparagraph{Virtual Haircut} 
By having an explicit strand with the shape texture $\mathbf{Z}_g$, we can trim its length and let the neural renderer infer how the hair would look like at the new length. In \reffig{fig:hair_manipulation}, we show that the UNet generalizes and produces realistic appearance even for this hair configuration that was never seen during training.


\mparagraph{Animation}
The explicit hair strands can also be deformed by slightly modifying the direction between adjacent strand nodes to convey a sense of dynamics of hair blowing in the wind. 
In~\reffig{fig:hair_manipulation}, we show examples of animating the hair with the appearance inferred for this novel hair configuration. 

\NEWNEW{
\mparagraph{Interpretable Strand Generator}
As the strands are generated from the latent space of a VAE, we can traverse this latent space to generate novel strands. 
In the supplemental video, we show that we can traverse each dimension of the latent space and discover interpretable controls for curliness, length, etc.}

\section{Limitations and Future Work}
\begingroup
\NEW{
Here we discuss several exciting future research directions to overcome current limitations of our method. 
First, for complicated hairstyles like the hair-bun in \reffig{fig:qualitative_compare}, it is challenging to infer the exact topology since most of it is occluded. Stronger priors for hairstyles learned from various subjects could help alleviate these issues. 
Second, although our hair model is fully editable, due to the complicated light transport, the generated appearance may not generalize to large hair movement, since lighting effects, like shadows, may be baked in the learned appearance. We will take it as future work to explore more physics-aware rendering since strand-level geometry is available from our model.  

}
\endgroup





\bibliographystyle{eccv_template/splncs04}
\bibliography{base}

\end{document}